\documentclass[letterpaper, 10 pt, conference]{ieeeconf}  

\IEEEoverridecommandlockouts                              

\usepackage{graphicx}
\usepackage{epstopdf}
\usepackage{multicol}
\usepackage{stfloats}
\usepackage{amsfonts}

\usepackage{mathrsfs}
\usepackage{multicol}
\usepackage{stfloats}
\usepackage{cite}
\usepackage{amsfonts}
\usepackage{mathrsfs}
\usepackage{amsfonts}
\usepackage{empheq}
\usepackage[linesnumbered,ruled,lined,boxed]{algorithm2e}
\usepackage[procnames]{listings}
\usepackage{algpseudocode}
\usepackage{color}
\usepackage[ruled]{algorithm2e}
\usepackage{fancybox}
\usepackage{framed}
\usepackage{amsfonts, color}
\usepackage{color}
\usepackage{setspace}
\usepackage[dvips]{epsfig}
\usepackage{framed}
\usepackage{psfrag, amssymb, amsmath, cite}
\usepackage[colorlinks,linkcolor=red,anchorcolor=blue,citecolor=blue]{hyperref}
\DontPrintSemicolon

\usepackage{verbatim}
\usepackage{graphicx, subfigure}

\usepackage{url}

\title{\LARGE \bf
2.5D Image based Robotic Grasping
}

\author{Yaoxian Song$^{1}$ $^{2}$, Chun Cheng$^{1}$ $^{2}$, Yuejiao Fei$^{1}$, Xiangqing Li$^{1}$, Changbin Yu$^{1}$
	\thanks{$^{1}$The authors are with School of Engineering at Westake University Emails: 
		{\tt\small \{songyaoxian, chengchun, feiyuejiao, lixiangqing, yuchangbin\} @westlake.edu.cn}}%
	\thanks{$^{2}$The authors are also PhD students in Computer Science at Fudan University.}
}

\begin{document}

\maketitle
\thispagestyle{empty}
\pagestyle{empty}

\begin{abstract}
We consider the problem of robotic grasping using depth + RGB information sampling from a real sensor. we design an encoder-decoder neural network to predict grasp policy in real time. This method can fuse the advantage of depth image and RGB image at the same time and is robust for grasp and observation height.
We evaluate our method in a physical robotic system and propose an open-loop algorithm to realize robotic grasp operation. We analyze the result of experiment from multi-perspective and the result shows that our method is competitive with the state-of-the-art in grasp performance, real-time and model size. The video is available in \url{https://youtu.be/Wxw_r5a8qV0}.
\end{abstract}

\section{Introduction}
Robotic grasping in picking and transferring one or more targets from a specific place to the other in unstructured environment is a fundamental problem in robotic control. The robot needs to perceive environment from multi modals before making decisions. Vision-based robotic grasping method has become  most popular approach.

Robotic grasping has been researched for decades and different researchers use different ways to solve this problem\cite{bicchi2000robotic}. Learning-based method is one of the mainstream research method of robotic grasping and has achieved impressive performance in the recent years. \cite{mahler2017dex}\cite{morrison2018closing} use deep learning to estimate the grasp policy with human labeled dataset or synthetic CAD dataset. The performance of deep learning method has a great improvement. \cite{zeng2018learning}\cite{thomas2018learning} realize high-level grasping operations by reinforcement learning.

Most existing methods  train the neural network including Convolutional Neural
Networks (CNNs) and Fully Convolutional Networks (FCNs) only using depth image or RGB image to output grasp policy individually. The dataset of the robotic grasping is very small comparing to computer vision problem. It is necessary to more extract information from the limited grasping dataset.  

In this paper, we propose a FCN grasping network (UG-Net V2) based on the U-Net\cite{ronneberger2015u} which generates an optimal grasping policy in pixel-wise. Our work is based on previous work UG-Net (\url{https://youtu.be/LJJRqmpYl2c}) which only using depth image to predict grasp policy. Because the depth sampling is unstable from different observation height but non-sensitive to the shadow and RGB image is non-sensitive to the height but exists shadow from the natural light, we propose UG-Net V2 to fuse advantages of both depth image and RGB image by using depth image and RGB image directly.

Our main contributions in this paper are: (1) We design UG-Net V2 which inputs depth and RGB images directly and outputs the grasp prediction in pixel-wise.  Some preprocessing is given to improve the result. (2) We use a physical experiment to evaluate the effectiveness of our method. The experiment results show that UG-Net V2 is competitive with the state-of-the-art in grasp performance, real-time and model size.

\section{Related work}
Grasping is a fundamental operation in robotic manipulation which has been researched for decades. Traditional methods\cite{siciliano2016springer} for the grasping concentrate on the analysis which are dependent on the target precise physical models and multiple properties like geometry, dynamic, friction, etc. Another popular method is learning-based. Many researchers design grasping algorithms using data-based and behavior-based learning methods.
\cite{morrison2018closing}\cite{lenz2015deep}\cite{detry2013learning}\cite{kappler2015leveraging} use deep learning to generate grasp policy. \cite{zeng2018learning}\cite{thomas2018learning}\cite{balasubramanian2012physical}\cite{levine2018learning}
\cite{sadeghi2018sim2real} explore the robotic task environment and generate the grasp policy by reinforcement learn. 

Robotic manipulation has been researched using multiple sensors. Current work mainly concerns about computer vision data. FCNs and CNNs have been applied successfully to the grasp prediction   \cite{mahler2017dex}\cite{lenz2015deep}\cite{arruda2016active}\cite{Pas2017Grasp} based on 2D or 3D (including 2D+depth image or pointcloud) computer vision. Furthermore, heterogeneous sensor modalities are popular these years. Many works try to fuse multi-modal data like vision, range, haptic data as well as language to let robot grasp target objects more stable and flexible\cite{calandra2018more}\cite{lee2018making}\cite{calandra2018more}\cite{zeng2019tossingbot}.

\section{Preliminary and problem formulation} \label{Sec:preliminary}
In this section, we study vision-based robotic grasping problem using parallel-jaw and RGBD image. A robust parallel-jaw is programmed to grasp a novel object placed on the table surface navigating by a fixed RGBD camera overhead. We need to learn a function map from the image space to the grasp space which we will define  later.    

\subsection{Problem definition}\label{Sec:problem_define}   
\textbf{Grasp definition} In this paper, robotic grasp is to predict a 5D grasp representation for single or multiple objects based on RGB and depth images. The 5D representation consists of position of the grasp $(x, y, z)$, orientation of the gripper $\phi$ and open width of gripper $w$. The representation in Cartesian coordinates can be defined as follows:
\begin{align}\label{eq:grasp_def}
	g=(x,y,z,\phi, w)
\end{align} 

\textbf{Transformation} In real grasp pipeline, we need to consider 5D grasp representation in different coordination. Here, we discuss grasp representation in image space and camera space. We assume the intrinsic and extrinsic parameter of the camera and physical properties of the robot are known.

The RGB and depth image taken from RGBD camera is $I = \mathbb{R}^{4\times H\times W}$ where $H$ is the image height and $W$ is the image width. In image space, the 5D grasp representation can be rewritten as:
\begin{align}
{\widetilde g} = (u,v,\widetilde\phi,\widetilde\omega)
\end{align}
where $(u,v)$ is the position of grasp in image coordination. $\widetilde\phi$ and $\widetilde\omega$ correspond the $\phi$ and $w$ in Cartesian coordinates. We can get the grasp representation in the robot base coordinates from following the Eq. \eqref{eq:convert2Cartesian}. 

\begin{align}\label{eq:convert2Cartesian}
g = {}_{Camera}^{Robot}T \times {}_{Image}^{Camera}T \times \widetilde{g}
\end{align}
where ${}_{Image}^{Camera}T$ is the transformation matrix from image space to camera space based on intrinsic parameter of the camera. ${}_{Camera}^{Robot}T$ is the transformation matrix from camera coordinates to robot base coordinates based on the extrinsic parameter of the camera.

\subsection{Objective}  \label{objective}
In the grasp prediction, we estimate each pixel's probability of the grasp instead of predicting the position of grasp directly. So we can redefine the grasp as
\begin{align}
	G = (Q, \widetilde{\Phi}, \widetilde{W})
\end{align}
where $Q, \widetilde{\Phi}, \widetilde{W} \in \mathbb{R}^{ H\times W}$ is each pixel's probability, orientation and gripper's width of the grasp respectively. We get the position of grasp $(u, v)$ by selecting the maximum probability of pixel position.

Instead of detecting the object and planning grasp, we try to get a grasp policy by pixel-wise metric on RGBD image. We define the a function M from the image input to grasp space:
\begin{align}
	G = M(I)
\end{align}
where $I \in \mathbb{R}^{4\times H\times W}$ denotes a RGBD image.

Our goal is to find a robust function $M_\theta$:
\begin{align}
\theta  = \mathop {\arg \min }\limits_\theta  \mathcal{L}\left( {G,{M_\theta(I) }} \right)
\end{align}
where $\mathcal{L}$ is the loss function between ground truth and $M_\theta$, $\theta$ is the parameter of function $M$. After the modeling process, we get the most robust function $M$ and the  optimal grasp $\widetilde g^* = \mathop {\max }\limits_Q G$ in camera space. Finally, we can get 5D grasp representation in  robot based coordinates via Eq. \eqref{eq:convert2Cartesian}.

\section{Method}\label{sec:constraints_formulation}
In this section, we will introduce a deep encoder-decoder fully convolutional network to approximate the function $M: I \to G$ defined in last section. We try to use supervised learning to learn a function $M_\theta(I)$ where $\theta$ is the weight parameter of the neural network and $I$ is the input RGBD image of the network. 

\subsection{Dataset generation}\label{sec:data_generation}
We use Cornell grasp detection dataset\cite{lenz2015deep} to train our network. This dataset is a small human-labeled dataset containing 1035 RGBD images of 280 different objects with ground truth labels of positive graspable rectangle and negative non-graspable rectangles. We project point cloud data into a depth image, concatenate with RGB image as $I \in \mathbb{R}^{4\times H\times W}$ and resize $I$ into $304 \times 304$. Then we augment the dataset like most of supervised methods by rotating, translating, cropping and scaling the raw data. 

\subsection{Raw data processing}\label{sec:constraint_kinematics}
We defined 5D grasp representation in \ref{Sec:problem_define} and here we will illustrate the raw data processing based on the grasp definition. In Cornell grasp detection dataset, antipodal grasp candidates are labeled with rectangles in image space. We choose the middle third of the rectangle as our grasp mask like \cite{morrison2018closing}.

\textbf{Grasp probability}
We use the new grasp mask as our train label and choose the center point of the mask as the position of grasping. We make a sparse binary label image in which positive pixel point is assigned as 1 and other pixel point as 0.

\textbf{Gripper orientation}
We define the gripper's orientation angle $\widetilde\phi$ in the range of $[-\frac{\pi}{2},\frac{\pi}{2}]$ and represent $\widetilde\phi$ as a vector $(\cos(2\widetilde\phi), \sin(2\widetilde\phi))$ on a unit circle of which value is a continuous distribution in $[-1, +1]$. \cite{hara2017designing} shows this processing is easy for the training.

\textbf{Gripper width}
We compute the gripper's open width in image space and the value $\widetilde w$ equals to each rectangle's width. We normalize the value by $\frac{1}{150}$ because the maximum width in the dataset is $150$ pixels. Furthermore, the gripper in robot base coordinates (physical world) can be converted by camera parameter and the depth value.

\textbf{Image input}
As discussed in \ref{sec:data_generation}, we use RGB image and depth image at the same time by concatenating the two image into a RGBD image $I \in \mathbb{R}^{4\times H\times W}$. The depth image is inpainted with OpenCV \cite{bradski2000opencv}. Because the dataset is sampled from camera which is real data with noise, we do not need to consider the noise filtering. RGB pixel value's scale is in $[0,255]$ and depth value's scale is $[20, 120]$ which is a typical value for Realsense SR300 and it indicates that two images have different scales. We scale the data value between $0$ and $1$ by min-max normalization following Eq. \eqref{eq:normalization}:

\begin{align}\label{eq:normalization}
	X^\star = \frac{X-min}{max - min}
\end{align}

\subsection{Learning network design} \label{sec:network}
\begin{figure*}
	\centering
	\includegraphics[width=0.8\textwidth]{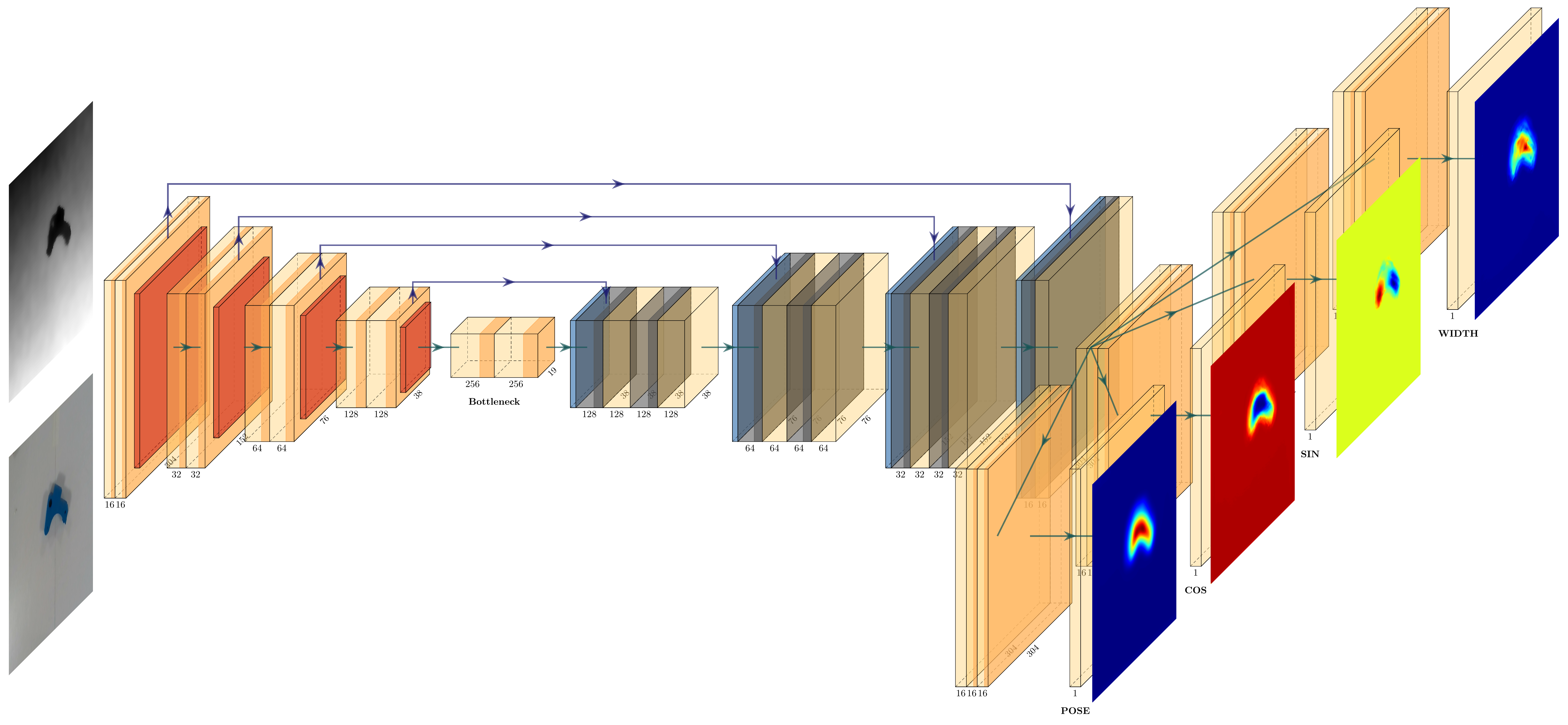} 
	\caption{UG-Net V2 Architecture. UG-Net V2 takes an inpainted depth image and RGB image as input. We use the backbone of U-net to extract features and reduce the number of channels for each layer. Outputs are three single channel images visualized in the form of heatmaps.} 
	\label{UG-Net_V2}
\end{figure*}

The architecture of UG-Net V2 is illustrated in Fig \ref{UG-Net_V2}. In this work, we design a fully convolutional network to approximate the function $M_\theta$. UG-Net V2 is a extended version of UG-Net based on U-net\cite{ronneberger2015u}. U-net is a typical multi-scale network and has the same size between input and output image. The network inputs an RGBD image $I$ and outputs normalized grasp representation by four output branches which predict probability $Q$, orientation vector $(\cos(2\widetilde\Phi), \sin(2\widetilde\Phi))$ and gripper's width $\widetilde{W}$ respectively. Function UG-Net V2  can fit $M_\theta(I) = (Q_\theta, \widetilde{\Phi}, \widetilde{W}) $ with $304 \times 304$ resolution. The orientation can also be computed with following Eq. \eqref{eq:arctan}. Our network has 2,327,876 (approximately 28MB) parameters and realize real-time running on our platform which we will introduce in the next section. 
\begin{align}\label{eq:arctan}
	\widetilde\Phi_\theta = \frac{1}{2}arctan\frac{sin(2\widetilde\Phi_\theta)}{cos(2\widetilde\Phi_\theta)}
\end{align}

\subsection{Proposed loss function and Training}
We use 80\% of our dataset for training and 20\% of the dataset for evaluation. We use \textit{Relu} activation function for all the layers except the last layer as \textit{linear}. We use MSE loss function Eq. \eqref{eq:loss_function} to train our UG-Net V2 10 epoches. The output is $(Q_\theta, (\cos(2\widetilde\Phi)_\theta, \sin(2\widetilde\Phi))_\theta, \widetilde{W}_\theta)$ and ground truth is $(Q_g, (\cos(2\widetilde\Phi)_g, \sin(2\widetilde\Phi))_g, \widetilde{W}_g)$. It takes about 100 minutes with bachsize 4 using two NVIDIA GTX 1080TI graphic cards.

\begin{align}\label{eq:loss_function}
\begin{array}{l}
\mathcal{L} = \frac{{{\lambda _q}}}{n}{\left\| {{Q_\theta } - Q_g} \right\|^2} + \frac{{{\lambda _{\cos }}}}{n}{\left\| {{\cos(2\widetilde\Phi)_\theta } - \cos(2\widetilde\Phi)_g} \right\|^2}\\
\\
+ \frac{{{\lambda _{\sin }}}}{n}{\left\| {{\sin(2\widetilde\Phi))_\theta } - \sin(2\widetilde\Phi))_g} \right\|^2} + \frac{{{\lambda _w}}}{n}{\left\| {{\widetilde{W}_\theta } - \widetilde{W}_g} \right\|^2}
\end{array}
\end{align}
where $\lambda _q, \lambda _{\cos} , \lambda _{\sin}, \lambda _w$ are the weight coefficients and we set all of default value as $1$.

\subsection{Grasping Metric}
We consider three performance metrics for the robotic grasp including success rate, robust grasp rate and planning time.

\textbf{Success Rate} The  percentage of successful grasps in all the grasp try.

\textbf{Robust Grasp Rate} The ratio of probabilities higher than 50\% of successful grasps through all the runs.

\textbf{Planning Time} The time consumed between receiving the raw data from the camera and UG-Net V2 outputting the 5D grasp representation prediction. Because our algorithm runs on the physical system including table environment or mobile platform, it is necessary to consider the system's real-time capability.

\section{Experiment}
In this section, we evaluate our UG-Net V2 in the physical environment. We choose a general benchmark including 3D-printed adversarial objects and household objects. Kinova Jaco 7DOF single arm robot is used to execute the grasp operation. 

\subsection{Assumption}
To realize our method, we need to finish some prefixed work. First, the RGBD image's coordinates should align with the depth image's. Second, we should synchronize the time stamp between RGB image topic and depth image topic. These two processes are easy to realize in simulation environment but need some engineering work in the physical system.

\subsection{Experimental Components}
To evaluate the performance of UG-Net V2, we test both in the simulation environment and physical. We establish a Kinova Jaco simulation environment using Pybullet\cite{coumans2016pybullet}. But in this paper we only give the physical experiment result and the simulation environment code is available in \url{https://github.com/aaronhd/pybullet_kinova7.git}. We use Intel RealSense SR300 RGBD camera to get the image information which is mounted on the wrist of the robot. All the computation is finished on a PC running Ubuntu16.04 with a Intel Core i7-8700K CPU and two NVIDIA Geforce GTX 1080ti graphic cards. It is noticed that we use two cards to accelerate training process and just use one card to running trained network.

\subsection{Test Objects}\label{sub_object}
\begin{figure}
	\centering
	\includegraphics[width=1.6in]{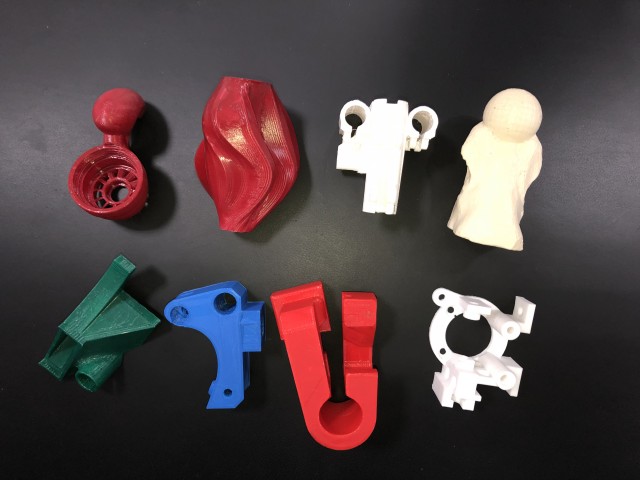}
	\caption{Test Objects}
	\label{Test_Object}
\end{figure}

We use eight adversarial objects to form our test set. The test set is shown in Fig \ref{Test_Object} which is proposed in Dex-Net 2.0 \cite{mahler2017dex}. These objects contain complex geometric features and less projection section information from some perspectives.

\subsection{Pipeline}
We design a deep encoder-decoder fully convolutional network using RGBD information to realize robotic grasping. We also train a same network only inputting depth information as baseline.

We design an open-loop controller to grasp target object from three different heights. During the experiment, although the sensor's producer declares the sensor can be used within a wide range, we find the depth raw data has much noise and is too sparse in different height. We also find RGB data is stable independent of the measuring distance. Different heights experiment results can indicate the effect of RGB information. The whole pipeline is presented in Algorithm \ref{algorithm:controller}.

\subsection{Result and Analysis}

\begin{figure}
	\centering
	\includegraphics[width=1.6in]{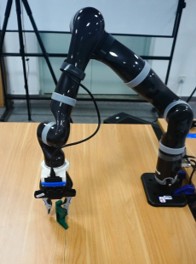}
	\caption{Robotic Setup}
	\label{set_up}
\end{figure}

\begin{figure}
	\centering
	\includegraphics[width=0.5\textwidth]{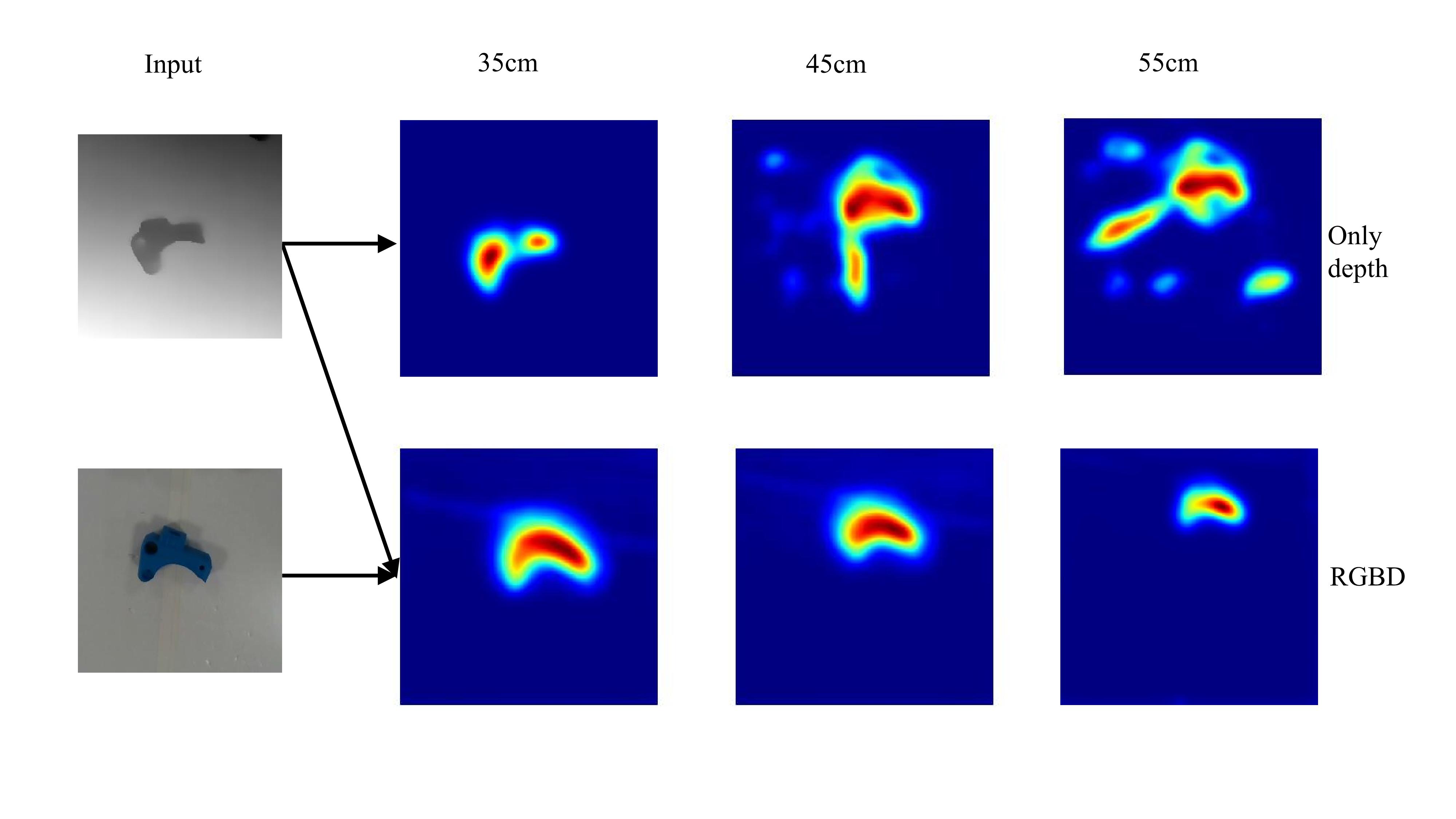}
	\caption{Grasp probability in pixel-wise}
	\label{grasp_visulization}
\end{figure}

\begin{figure}
	\centering
	\includegraphics[width=0.5\textwidth]{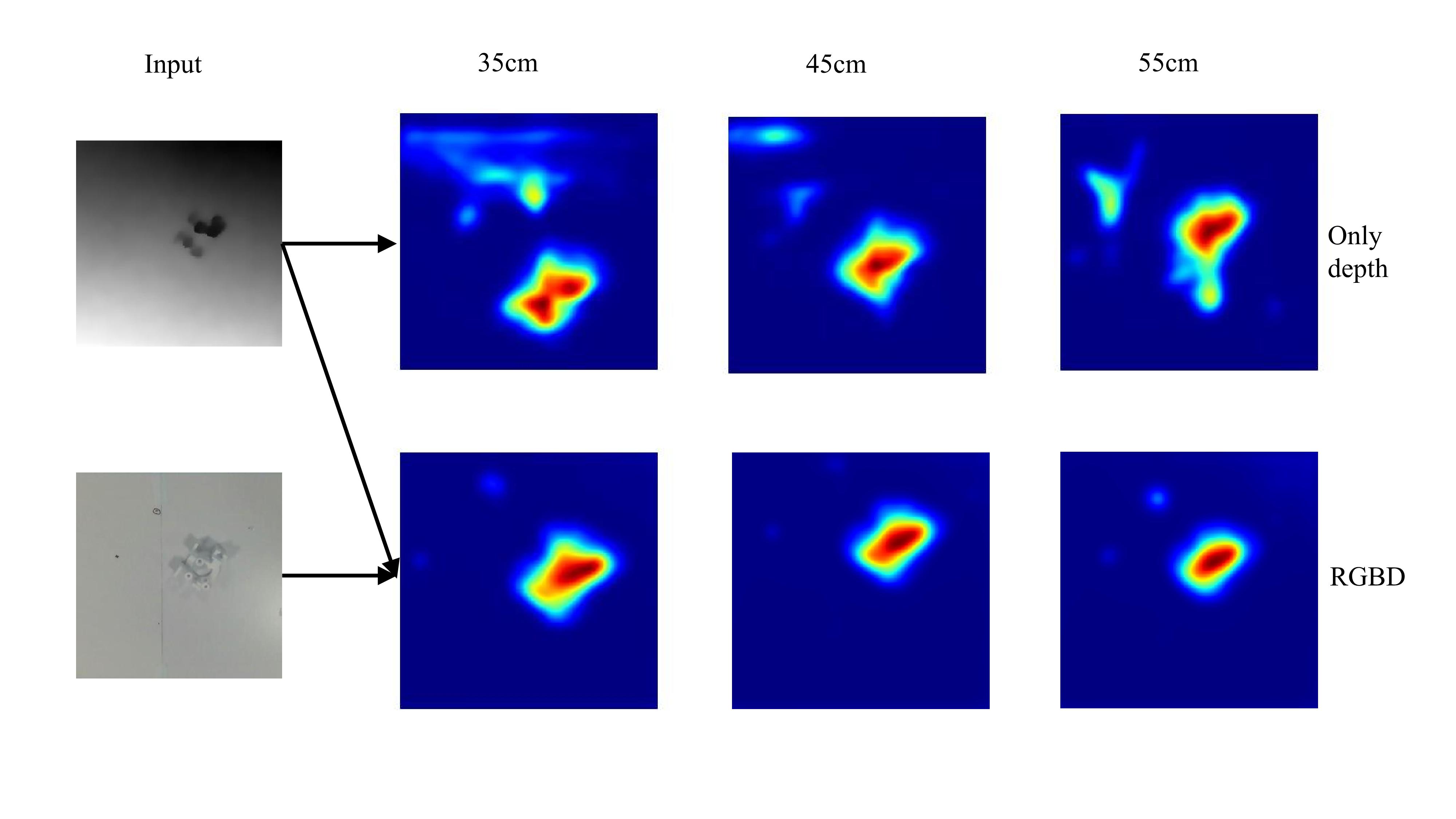}
	\caption{Grasp probability in pixel-wise with depth data mission}
	\label{grasp_missing}
\end{figure}

In the experiment, we perform grasp trails on the adversarial objects talked in \ref{sub_object} for more than 480 times. Our robot grasps each object 10 trails individually with the set-up in Fig. \ref{set_up} in three observation heights (35, 45, 55 centimeter). We use the depth-only network as the baseline to realize robotic grasping and evaluate the performance of depth confused with RGB data method. The experiment result is shown in TABLE \ref{baseline_comparision}. We can see that the method fused RGB and depth data together achieves much improvement in grasp performance. With both depth-only and RGB+depth, we can see that the robust grasp rate is always lower than success rate. Because we get the optimal grasp by selecting the highest probability point in pixel-wise talked in \ref{objective}, the grasp position probability maybe the optimal grasp but lower than 50\%. The sensitivity to the observation height shows that the smaller distance between the camera and the object can achieve better grasp performance in only depth scenario and the bigger distance between the camera and the object grasp better in RGB+depth scenario. We believe depth data sampling from the sensor will be sparser, noisier or null with the height increasing which will badly affect the grasp performance. However, RGB data sampling is independent of the height within specific measurement range and the higher observation height can get more global information which is used to predict the grasp policy. Fig. \ref{grasp_visulization} and Fig. \ref{grasp_missing} show the grasp probability prediction visualization which also indicate the advantage of RGB information for the grasp prediction.

\begin{table*}[]
	\centering
	\begin{tabular}{ccccc}
		\hline
		\textbf{\begin{tabular}[c]{@{}c@{}}Obersvation Height\\ (cm)\end{tabular}} & \textbf{\begin{tabular}[c]{@{}c@{}}Depth\\ Success Rate\end{tabular}} & \textbf{\begin{tabular}[c]{@{}c@{}}Depth\\ Robust Grasp Rate\end{tabular}} & \textbf{\begin{tabular}[c]{@{}c@{}}Depth+RGB\\ Success Rate\end{tabular}} & \textbf{\begin{tabular}[c]{@{}c@{}}Depth+RGB\\ Robust Grasp Rate\end{tabular}} \\ \hline
		\multicolumn{1}{c|}{35}                                                    & 83.34\%                                                               & 78.31\%                                                                    & \textbf{96.39\%}                                                          & \textbf{90.36\%}                                                               \\ \hline
		\multicolumn{1}{c|}{45}                                                    & 86.59\%                                                               & 78.05\%                                                                    & 94.05\%                                                                   & 82.14\%                                                                        \\ \hline
		\multicolumn{1}{c|}{55}                                                    & \textbf{86.67\%}                                                      & \textbf{82.22\%}                                                           & 93.33\%                                                                   & 75.56\%                                                                        \\ \hline
	\end{tabular}
	\caption{Results from different observation heights with \textbf{only depth} and \textbf{depth+RGB} methods.}
	\label{baseline_comparision}
\end{table*}

\begin{algorithm}[t!]
Initialize home pose of the robotic gripper $P_{home}$;\;
Load the UG\_Net V2 model parameter $M_\theta$;\;

\BlankLine
\While{Running}{
	\BlankLine
	
	Get a frame of the RGB image $I_{RGB}$ and depth data $I_{depth}$ in the form of matrix.\;
	Preprocess the data: convert RGB image into grey image $I_{grey}$, inpaint the depth image. \;
	Concantenate  $I_{grey}$ and  $I_{depth}$ into a matrix $304 \times 304 \times 2$: $I_{input}$. \;
	Input the $I_{input}$ to the $M_\theta$ to get the grasp prediction  $G$. \;
	Obtain the maximum probability position $\widetilde{g}$ in pixel space. \;
	Convert $\widetilde{g}$ to $g$ according to  Eq. \eqref{eq:convert2Cartesian}. \;
	\BlankLine
	\uIf{$g$ is out of range of the robotic arm work space OR the $z$ predicition is not higher than table surface}{
		The predicition is invalid and need to re-predict grasp $G$. \;
	}
	\Else{
		Move the gripper to the target position and grasp the object. \;
		Move the gripper to the home pose $P_{home}$. \;
	}
	
	\uIf{the gripper open is completely closed}{
		Grasp failed.
	}
	\Else{
		Grasp successed. \;	
		Release the gripper. \;
	}
}
\caption{Grasping based on UG-Net V2: raw data processing, policy prediction, execution}
\label{algorithm:controller}
\end{algorithm}

\subsection{Some Comparison}
We also give some comparison with other work which is shown in TABLE \ref{comparision_other_work}. Although the robotic grasping task is a complex operation for different scenarios and does not have similar standard, we still manage to compare our work with some existing works using same benchmarks. It need to be noticed that GQ-CNN\cite{mahler2017dex} and GG-CNN\cite{morrison2018closing} used GTX 1080 and GTX 1070 NVIDIA graphic card respectively. Our work uses one GTX1080ti NVIDIA graphic card to realize grasp prediction and only takes 8ms for the prediction. Our hardware is the best of the three, but it is a general capacity of calculation currently. According to above-mentioned situation, our work  is still competitive. The success rate and the robust grasp rate of our work are both better than other two existing works. 

\begin{table*}[]
	\centering
	\begin{tabular}{lcccc}
		\hline
		& \textbf{GQ-CNN} & \textbf{GG-CNN} & \multicolumn{1}{l}{\textbf{\begin{tabular}[c]{@{}l@{}}UG-Net V2\\ Baseline\end{tabular}}} & \multicolumn{1}{l}{\textbf{UG-Net V2}} \\ \hline
		\multicolumn{1}{l|}{\textbf{Success Rate(\%)}}      & 80              & 84              & 85.88                                                                                     & \textbf{94.55}                         \\ \hline
		\multicolumn{1}{l|}{\textbf{Robust Grasp Rate(\%)}} & 43              &                 & 79.61                                                                                     & \textbf{82.49}                         \\ \hline
		\multicolumn{1}{l|}{\textbf{Planning Time(ms)}}     & 800             & 19              & \textbf{8}                                                                                         & \textbf{8}                                      \\ \hline
		\multicolumn{1}{l|}{\textbf{Model Size(Approx.)}}   & 75MB            & \textbf{0.5MB}           & 28MB                                                                                      & 28MB                                   \\ \hline
	\end{tabular}
	\caption{Comparison with other existing works, our method is competitive in the performance.}
\label{comparision_other_work}
\end{table*}

\section{Conclusions} \label{sec:conclusions}
In this paper, we propose an encoder-decoder neural network to predict grasp policy using 2.5D image information. The depth image can provide physical scale geometric features for the grasp policy generation and the RGB image can provide diverse robust information in camera space. We fuse the two modal data to predict  grasp policy more robustly. The experiment result shows that our network structure can predict the grasp probability more stable and robust in Fig. \ref{grasp_visulization} and Fig. \ref{grasp_missing}. We design an open-loop controller to evaluate the grasp performance in the physical system showed in Algorithm \ref{algorithm:controller} and achieves high grasp success rate comparing to our baseline (only depth). We also compare some the state-of-the-art and our method is competitive in grasp performance, real-time and model size. We believe our method is very great for mobile platform deployment as a general module. Our experiment video is available in \url{https://youtu.be/Wxw_r5a8qV0}.

\bibliography{Feasibility}
\bibliographystyle{ieeetr}

\end{document}